# Web-Based Expert System for Civil Service Regulations:RCSES

Mofreh A. Hogo*
Electrical engineering Technology Dept.,
Higher Institution of Technology Benha,
Benha University, Egypt.

Khaled Fouad
Central Lab. for Agricultural
Expert Systems (CLAES)

Fouad Mousa
Business management Dept,
Faculty of commerce,
Assuit university

*the corresponding author, ___________________________

*Abstract—* **Internet and expert systems have offered new ways of sharing and distributing knowledge, but there is a lack of researches in the area of web-based expert systems. This paper introduces a development of a web-based expert system for the regulations of civil service in the Kingdom of Saudi Arabia named as RCSES. It is the first time to develop such system (application of civil service regulations) as well the development of it using web-based approach. The proposed system considers 17 regulations of the civil service system. The different phases of developing the RCSES system are presented, as knowledge acquiring and selection, ontology and knowledge representations using XML format. XML-Rule-based knowledge sources and the inference mechanisms were implemented using ASP.net technique. An interactive tool for entering the ontology and knowledge base, and the inferencing was built. It gives the ability to use, modify, update, and extend the existing knowledge base in an easy way. The knowledge was validated by experts in the domain of civil service regulations, and the proposed RCSES was tested, verified, and validated by different technical users and the developers' staff. The RCSES system is compared with other related web based expert systems, that comparison proved the goodness, usability, and high performance of RCSES.**

## I. INTRODUCTION

An expert system is a computer program that simulate the problem-solving behavior of a human, it is composed of a knowledge base (information, heuristics, etc.), inference engine (analyzes the knowledge base), and the end user interface (accepting inputs, generating outputs). The path that leads to the development of expert systems is different from that of conventional programming techniques. The concepts for expert system development come from the subject domain of artificial intelligence (AI), and require a departure from conventional computing practices and programming techniques. Expert systems (ES) emerged as a branch of artificial intelligence (AI), from the effort of researchers to develop computer programs that could reason as humans [1]. Many organizations have leveraged this technology to increase productivity and profits through better business decisions the work in [2-5] are one of most commercially successful branches of AI [6]. Although there have been reports of ES failures [7] and [8], surveys [9] and [10] show that many companies have remained enthusiastic proponents of the technology and continue to develop important and successful applications. Internet technology can change the way that an ES is developed and distributed. For the first time, knowledge on any subject can directly be delivered to users through a web based ES. Grove [11] provided some examples of web-based expert systems in industry, medicine, science and government and claimed that "there are now a large number of expert systems available on the Internet." He argued that there are several factors that make the Internet, by contrast to standalone platforms, an ideal base for KBS (knowledge based system) delivery. These factors include: The Internet is readily accessible, web-browsers provide a common multimedia interface, several Internet-compatible tools for KBS development are available, Internet-based applications are inherently portable, and emerging protocols support co-operation among KBS. He also identified several problems in the development of web-based KBS: Keeping up with rapid technological change to servers, interface components, inference engines, and various protocols; and reducing the potential delivery bottleneck caused by communication loads and a limited infrastructure Adams [12] pointed out that "there are numerous examples of expert systems on the web, but many of these systems are small, non-critical systems." The most successful example is probably the web-based legal expert system reported by Bodine [13], who remarked that "Law firms are collecting hundreds of thousands of dollars in subscription fees from clients who use their question-and-answer advisory services based on the web.". Contrary to Grove and Adams, Huntington [14] stated that "there are not many ES on the web" due to the fact that the Internet was not created with applications such as expert systems in mind. As a result, the manner in which the web and web browsers interface made it difficult to perform the actions required by ES. Athappilly [15] reported a "dynamic web-based knowledge system for prototype development for extended enterprise." He suggested that the use of emerging Internet technology made the development of multifunctional AI systems relatively easy and less expensive, but that many users in the business arena were unaware of these technologies and their potential benefits. Consequently, the business community was not aware of the value or educated to deploy the potential available from these technologies in making their business more efficient and competitive. The goals of the work presented in this paper are: *Firstly* to develop a web-based expert system equipped with an integrated knowledge for the regulations of the civil service system in Saudi Arabia, *secondly* building a knowledge base regarding the rules and regulations of the civil services based XML technique. *Thirdly* to develop an easy and attractive interface for the user to deal with the system for the introduction of inputs (entering ontology and knowledge base, queries or conditions) and view the results easily.

The overall structure of the proposed RCSES system consists of two main units the knowledge base builder unit and the RCSES interface unit. The knowledge base builder unit includes the ontology builder tool, morphological analyzer, and model builder tool. The expert system interface unit includes: data for the user selection, inference engine, the system results, and the reasoning engine. The final form of the proposed RCSES is presented too, with different web pages, the URL for the proposed work is: http://rcses.com.

The rest of the paper is organized as: Section 2 introduces the knowledge acquisition and selection. Section 3 presents the knowledge representation of the civil service regulations including the Ontology and the Knowledge Rules representation. Section 4 introduces the RCSES system development stage. Section 5 introduces the results (system usability and testing). Section 6 is reserved for the comparison with other related web based expert systems. Finally section 7 is reserved for the conclusion and future extensions.

## II.     KNOWLEDGE ACQUISITION AND SELECTION

A precise domain is required by an expert system, the domain must be compact and well organized. The quality of knowledge highly influences the quality of expert system. The knowledge base is the core component of any expert system; it contains knowledge acquired from the domain expert. Building the knowledge base with the help of domain expert is the responsibility of knowledge engineer. The first task of any expert system development is the knowledge acquisition; which is one of the most important phases in the expert system development life cycle. The process of knowledge acquisition is difficult especially in case if the knowledge engineer is unfamiliar with the domain. The goal of knowledge acquisition step is to obtain facts and rules from the domain expert so that the system can draw expert level conclusions. Knowledge acquisition is crucial for the success of an expert system and regarded as a bottleneck in the development of an expert system. The main reason for this bottleneck is communication difficulties between the knowledge engineer and the domain expert. Knowledge acquisition in the proposed RCSES was not a big problem; where the regulations of the civil service were selected from the web site of the ministry of the civil service then we use it as it is after the validation from the experts in the civil service domain.

The source of the regulations is: http://www.mcs.gov.sa. The system is developed based on 17 regulations of the civil service system in Saudi Arabia. More details can be obtained from the web site of the proposed web-based expert system: http://rcses.com.

## III.     THE KNOWLEDGE REPRESENTATION

After the step of the domain identification and knowledge acquiring from a participating expert of civil service regulation documents, a model for representing the knowledge must be developed. Numerous techniques for handling information in the knowledge-base are available; most expert systems utilize rule-based approaches. The knowledge engineer, working with the expert, must try to define the possible best structure. Other commonly used approaches include decision trees, blackboard systems and object oriented programming. Knowledge representation has been defined as "A set of syntactic and semantic conventions that make it possible to describe things. The syntax of a representation specifies a set of rules for combining symbols to form expressions in the representation language. The semantics of a representation specify how expressions so constructed should be interpreted (i.e. how meaning can be derived from a form). In the proposed system RCSES, the knowledge representation methodology uses XML format. Where, two elements of knowledge, ontology and model rules are represented using XML format. The overall knowledge structure is shown in Fig.1.

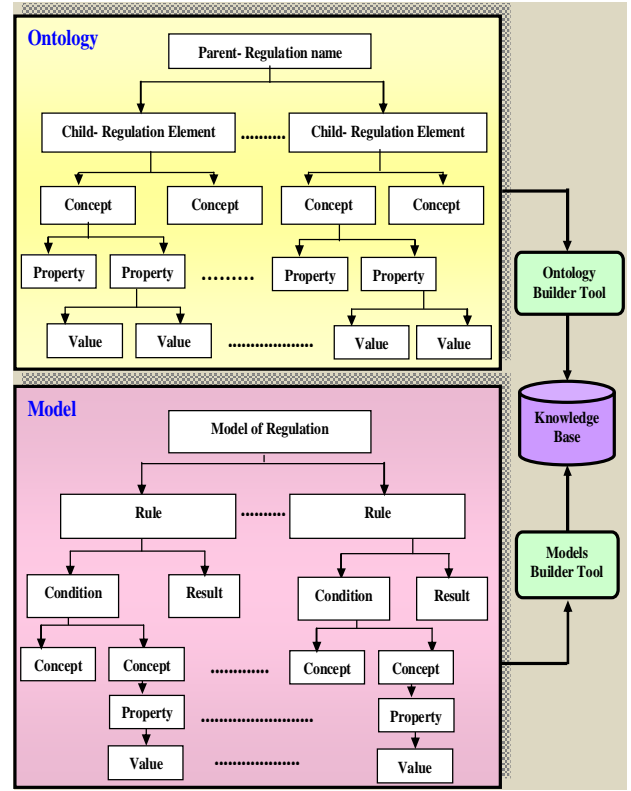

Figure 1.   Knowledge structure.

### A. Ontology Representation

An ontology is defined as ''formal, explicit specification of a shared conceptualization''. It represents the concepts and the relationships in a domain promoting interrelation with other models and automatic processing. Ontologies are considered to be a proper mechanism to encode information in modern knowledge-intensive applications, so they have become one of the most used knowledge representation formalism. They allow the enrichment of data with semantics, enabling automatic verification of data consistency and making easier knowledge base maintenance as well as the reuse of components. Ontology is a conceptualization of a domain into a human understandable, machine-readable format consisting of entries. The ontology entry has three layers: concepts, properties of concept and values of the properties.

*The proposed ontology structure* is illustrated by a real example for the ontology structure in the proposed RCSES as shown in Fig. 2; which describes the structure of separation regulation in civil service system using XML format. The interpretation of the example as following:

1. Node of "OntParent" represents regulation data and has attribute "ParentName" its value means the regulation name as: "التعيين في الوظائف العامة".

2. Child nodes "OntChild" represent the decision result for each part in regulation data and has attribute "ChildName" its value means the decision result of the regulation as: "التوظيف في وظائف المرتبة السادسة حتي العاشرة - مؤقت".

3. Node "OntConcept" represents the concepts in the regulation, and has attribute "ConceptName" its value is the concept names in the selected part in the regulation as:"الإعلان"

4. Node "OntVal" represents the values in the selected part in the regulation, it has an attribute "ValueName", which specifies the values of the OntVal, as "يوجد إعلان","لايوجد إعلان".

The proposed system developed a tool to build the domain ontology, which is easy to use and dynamic one for the acquired knowledge.

### B. Rules of Knowledge Representation

The knowledge can be formulated as shown in the following simple statements: *IF the 'traffic light' is green THEN* the action is go, as for example: *IF the 'traffic light' is red THEN the action is stop.* These statements represented in the IF-THEN form are called production rules or just rules. The term 'rule' in artificial intelligence, which is the most commonly type of knowledge representation, can be defined as IF-THEN

```
<KSA_Civil_Ontology>
<OntParent ParentName="التعيين في الوظائف العامة">
<OntChild ChildName="التوظيف في وظائف المرتبة السادسة حتي العاشرة - مؤقت">
<OntConcept ConceptName="الإعلان">
<OntVal ValueName="يوجد إعلان"/>
<OntVal ValueName="لا يوجد إعلان"/>
</OntConcept>
</OntChild>
<OntChild ChildName="التوظيف في وظائف المرتبة الرابعة عشرة فما فوق">
<OntConcept ConceptName="قرار من مجلس الوزراء">
<OntVal ValueName="يوجد قرار"/>
<OntVal ValueName="لا يوجد قرار"/>
</OntConcept>
</OntChild>
</OntParent>
</KSA_Civil_Ontology>
```

Figure 2.   Sample of ontology structure in XML format.

```
<KSA_Civil_Regulation>
<Model ModelName="إنهاء الخدمة">
<Rule Name="R1" RegItem="إنهاء الخدمة بالاستقالة" NoTrueFinding="0">
<Finding Cpt="تقديم الاستقالة وقبولها" Prop="Value" Val="الاستقالة"
Equal="Yes" ExistInWM="No"/>
</Rule>
<Rule Name="R2" RegItem="إنهاء الخدمة بطلب الإحالة على التقاعد"
NoTrueFinding="0">
<Finding Cpt="طلب الإحالة على التقاعد قبل بلوغ السن النظامية" Prop="Value"
Val="تقديم الطلب قبل بلوغ السن النظامية" Equal="Yes" ExistInWM="No"/>
</Rule>
</model>
</ KSA_Civil_Regulation>
```

Figure 3.   Sample of rules structure in XML format

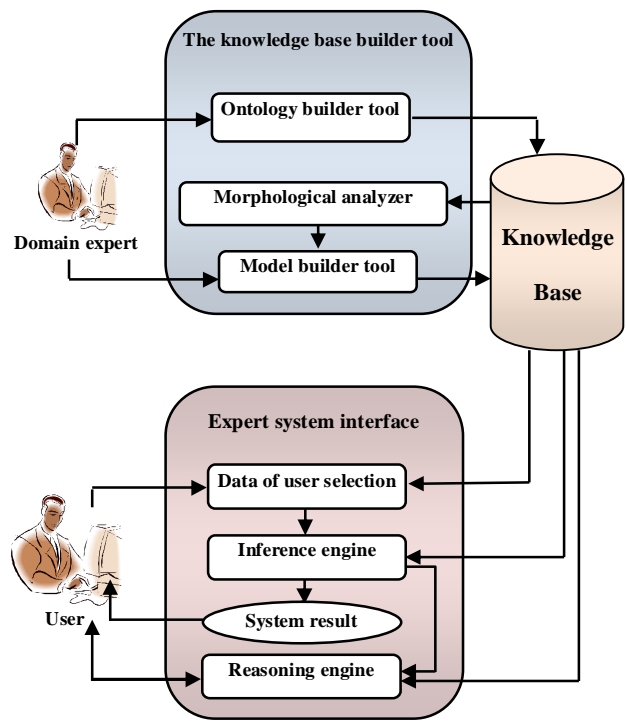

Figure 4.   Structure of the proposed system RCSES.

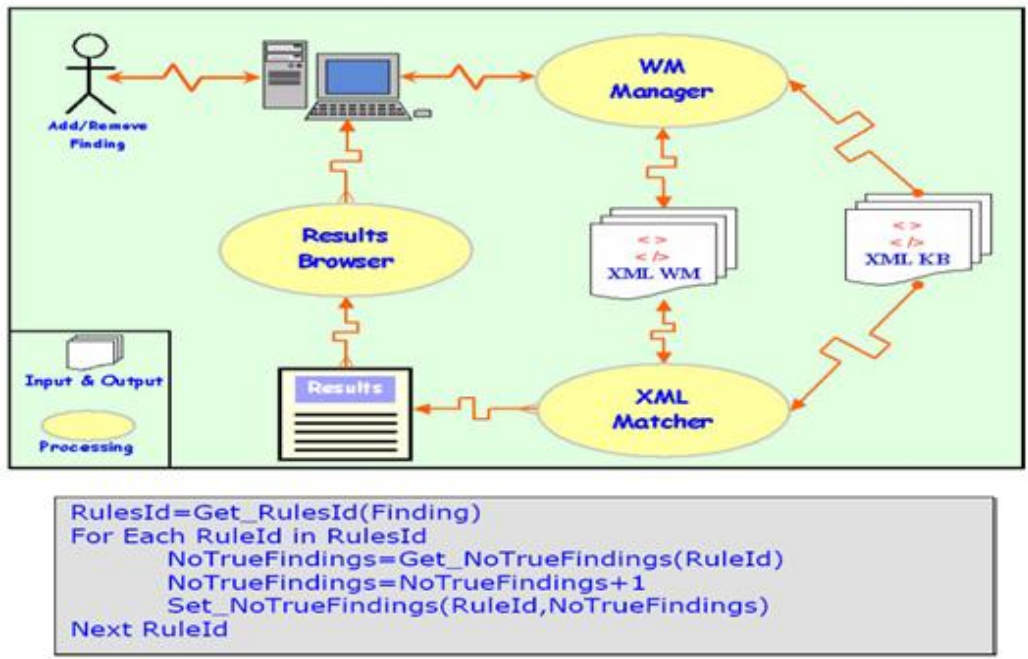

```
RulesId=Get_RulesId(Finding)
For Each RuleId in RulesId
        NoTrueFindings=Get_NoTrueFindings(RuleId)
        NoTrueFindings=NoTrueFindings+1
        Set_NoTrueFindings(RuleId,NoTrueFindings)
Next RuleId
```

Figure 5.   (a): Inference engine block diagram.

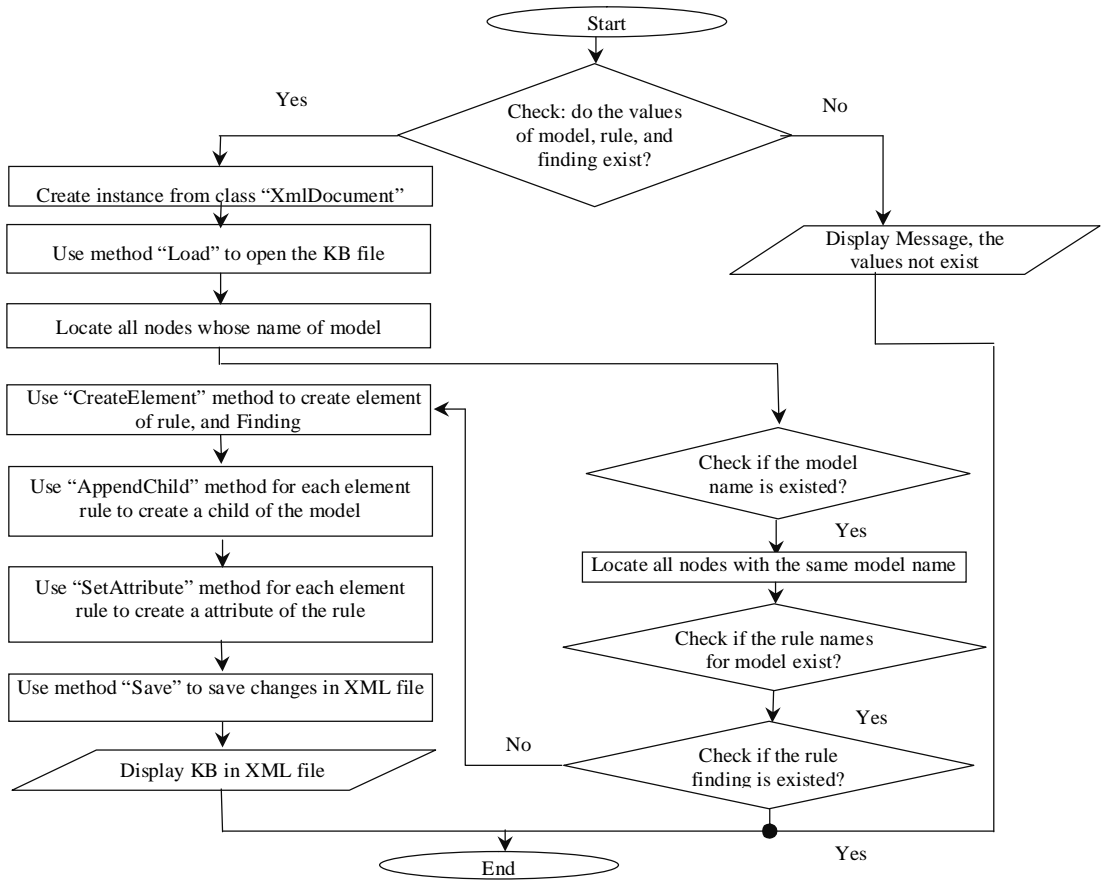

Figure 5.   (b): The algorithm for the inference engine.

structure that relates given information or facts in the IF part to some action in the THEN part. A rule provides some description of how to solve a problem. Rules are relatively easy to create and understand. Any rule consists of two parts: the IF part, called the antecedent (premise or condition) and the THEN part called the consequent (conclusion or action). The basic syntax of a rule is: *IF <antecedent>THEN <consequent>*. The rules in XML format have a different structure with the previous meaning but in different format. Sample of rule built in the proposed RCSES is shown in Fig.3; it can be interpreted as following:

1. <KSA_Civil_Regulation>; represents the total regulations' in the domain of the civil regulations.
2. The node of "Model" represents a regulation data and has attribute "ModelName" its value takes the regulation name as "إنهاء الخدمة".
3. The child nodes "Rule" represent the decision rule for each part in regulation; takes R1, R2,...and so on.
4. The node attribute "RegItem" represents the consequent in the selected rule in the regulation.
5. The node "Finding" represents antecedent in the rule.
6. The condition equal to attribute "Cpt" value as concept, attribute "Prop" value as property, and attribute "Val" value as value. The condition for example equal to separation "تقديم الاستقالة =الاستقالة وقبولها".

The proposed system RCSES, developed a tool component to build the domain model rules for the acquired knowledge in an efficient, easy, and dynamic way for changing and/or modifying the rules as required.

## IV. RCSES DEVELOPMENT

The development of the proposed RCSES system includes the development of different sub-systems. Fig.4 shows the block diagram of the entire proposed system RCSES.

The following section describes the different sub-systems in the proposed RCSES system as well as its functionality in the proposed system RCSES:

1. The knowledge builder unit: Developed to enable the expert or knowledge engineer to enter and save domain ontology and domain rules in an efficient and easy way.

2. The morphological analyzer unit: Developed to check if the entered words or sentences in the domain model rules are found in the ontology stored in the knowledge base or not; to avoid the unwanted reasoning results or the results concluded due to the user inputs errors and mistakes.

3. Web page interface of the proposed RCSES system: For using purposes includes entering ontology, knowledge, and the use of the expert system by entering the questions and get the output results with reasoning too.

4. The entire control and operation of the system is done by the inference engine; that is developed using Visual

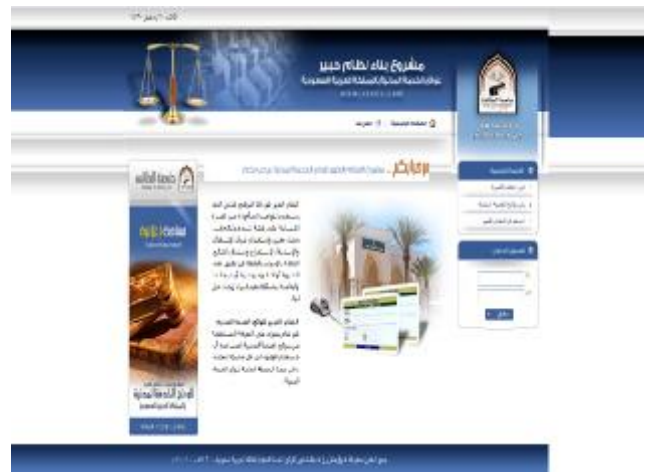

Figure 6. Home page of the proposed RCSES.

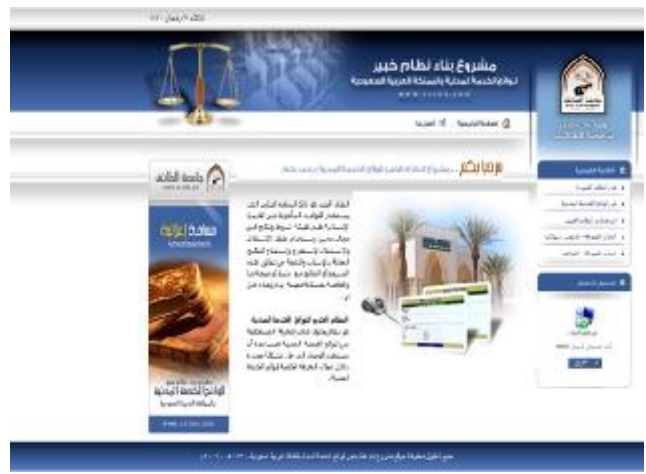

Figure 7. Web page for entering the ontology & knowledge base.

Basic dot net (VB.NET); which handles the knowledge in format of XML to get the result from the XML file (Knowledge base) that stores the knowledge rules. Fig. 5 shows the block diagram of inference engine. The main roles of the inference engine are summarized as: It applies the expert domain knowledge to what is known about the present situation to determine new information about the domain. The inference engine is the mechanism that connects the user inputs in the form of answers to the questions to the rules of knowledge base and further continues the session to come to conclusions. This process leads to the solution of the problem. The inference engine also identifies the rules of the knowledge base used to get decision from the system and also forms the decision tree.

### A. Reasoning Mechanism

The reasoning mechanism consists of three main components namely: working memory manager (WM manager), XML matcher, and result browser. Fig.5 (a) depicts the main components of the reasoning mechanism, while Fig.5 (b) illustrates the inference engine flowchart.

- *WM Manager Component:* The WM manager interacts with the user-friendly interface to get the concepts and its properties as well as the values of those properties by using communication model. The user-friendly interface permits the user to edit his complaints easily. This complaint is considered as a user finding. When the user selects concept, property, and value to be entered in the working memory, the WM manager creates an XSL query statement, which represents these findings.

- *XML Matcher Component:* In this methodology the rule is succeeded when all its child nodes are existed in working memory. This is achieved when the attribute 'ExistInWM'of every child node is set to "Yes". So, the matcher gets those succeeded rules by comparing the value of the attribute 'NoTrueFindings' of every parent node and the number of the child nodes in this rule and select the matched one. The succeeded rules are store in the result store for later use by display result component.

- *Result Browser Component:* Result Browser component gets the value of the attribute disorder for every rule in the result store, and pass value to the user interface by using communication model to display it to the user.

## V. RESULTS OF THE PROPOSED SYSTEM RCSES

The proposed RCSES was evaluated with different users, including "RCSES" developers' staff, and technical people. The system is validated by experts in the field of Civil service regulations. Tests of the system were carried out by the developers to make sure the system would work correctly as well as the RCSES is web based expert system, another validation and evaluation for the RCSES will be carried out through the using through the web and the feedbacks from the users will be considered for any comments and modifications. This output results from the developed RCSES will be interpreted via three parts as: The home page (http\\www. rcses.com) of the system, the web page for ontology building and knowledge building of the civil service regulations, and the web pages of the using the RCSES as shown in Figures 6,7,8, 9, and 10. Description of these units and sub-units of web pages are as following:

- The home web page is illustrated in Fig.6, which represents the user and expert interface to the RCSES system.

- The web page for the ontology and knowledge base builders is shown in Fig. 7; it assists the ability to the experts in entering the knowledge base ontology from the web page and the modification or extension of the knowledge base ontology.

- The web page for the ontology builder is shown in Fig. 8(a), while Fig. 8(b) shows the web page for the knowledge base builder; it assists the ability to enter the knowledge base from the web page and the modification or extension of the knowledge base.

- The web page for using the RCSES system that is shown in Fig. 9; it assists the ability to use the RCSES system, by entering the question (select the attributes, then selection the attributes values, continue until select all or the majority of the attributes then find the expected results and decisions. Finally you can see the reasoning; which can be viewed in form of HTML).

- Samples from the use of the RCSES and the corresponding output results, and for more details about the use of the RCSES system visit the rcses.com, we presented a Case of Employing in the governmental jobs التعيين في الوظائف العامة ", the steps for selecting the attributes and the conclusion inferred are illustrated through Figures 10(a..j); this case describes the all steps needed from the starting selecting the attributes, followed by the selection of the attributes vales, followed by the conclusion and the expected results, and the sure results. The system presented in this paper RCSES was not easy to be developed, but after the development it can be used by anyone with Internet access and a web browser easily. The web based ES made the evaluation and maintenance of "RCSES" easier than a conventional ES. There is no need to install the system in advance. It is easy to collect feedback. Visitors can be easily traced and analyzed, by collecting information, it was possible to profile the users and determine the value of the system. Compared with traditional ES development tools, the web design software simplifies the user interface design. XML-based user interfaces allow the incorporation of rich media elements. Hyperlinks provide an extra facility in enhancing ES explanation and help functions; users can access the relevant web site easily. The WWW helps in acquiring the knowledge needed in constructing the knowledge base. Any knowledge updating and maintenance can be handled centrally. Useful links are incorporated to help the user understand and interpret the ES recommendations. The e-mails, feedback forms and other Internet communication functions allow users to question and comment on the system.

## VI. COMPARISON WITH OTHER RELATED WBES

The comparisons with other similar web based expert systems are carried from different points of views as following:

1. KR: the Knowledge Representation,
2. KBR: Knowledge Base Repository,
3. KM: Knowledge Modification,
4. KB: Knowledge Builder,
5. IR: Inference Engine,
6. EP: Execution Platform,
7. PE: Programming Environment,
8. RT: Response Time for results seen.

From our searches of the literature and the presented comparison, there is no reporting on the topic, there is no any

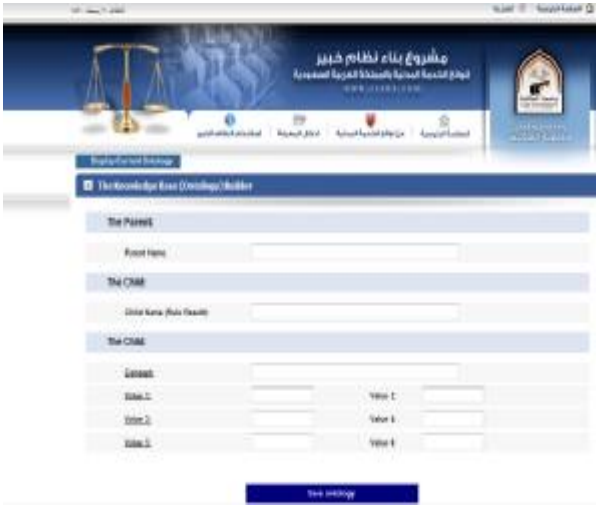

Figure 8. (a): Page for entering the ontology.

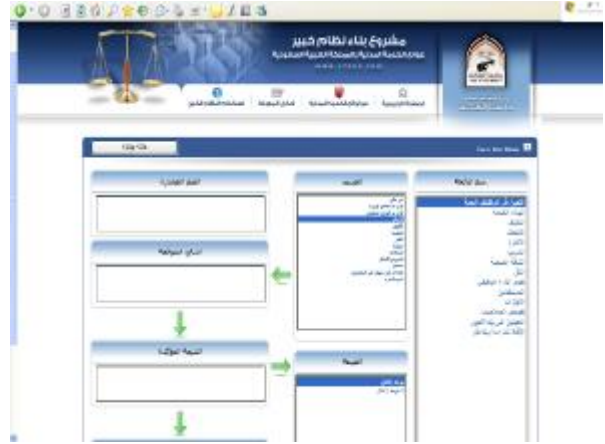

Figure 10. (a): Real example using RCSES step 1.

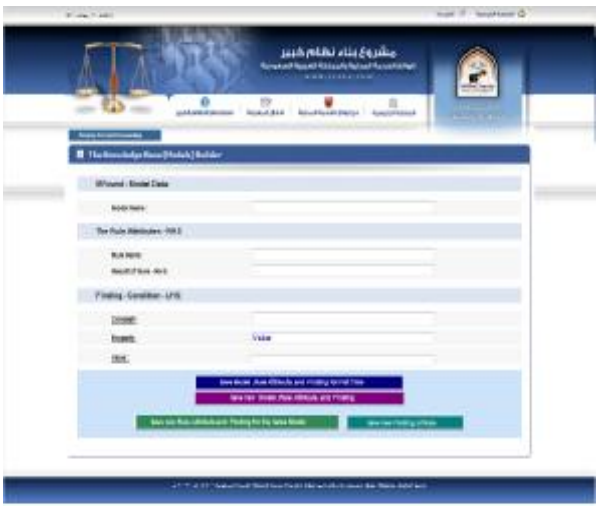

Figure 8. (b): Page for entering the Knowledge base.

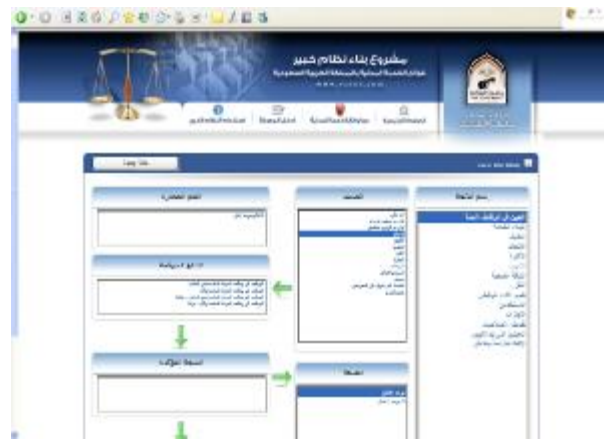

Figure 10. (b): Real example using RCSES step 2.

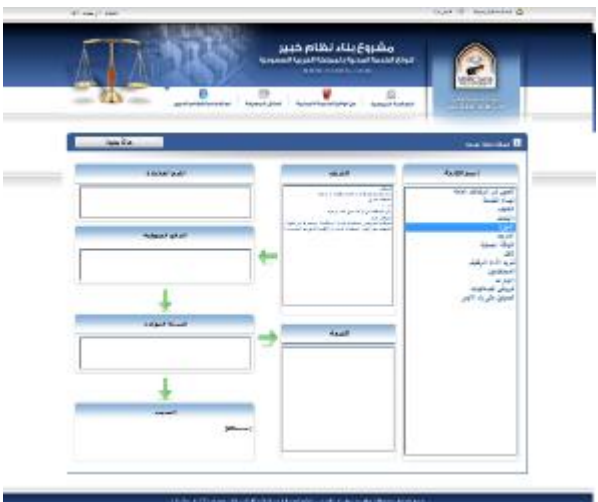

Figure 9. Using the RCSES system for questions and inferring.

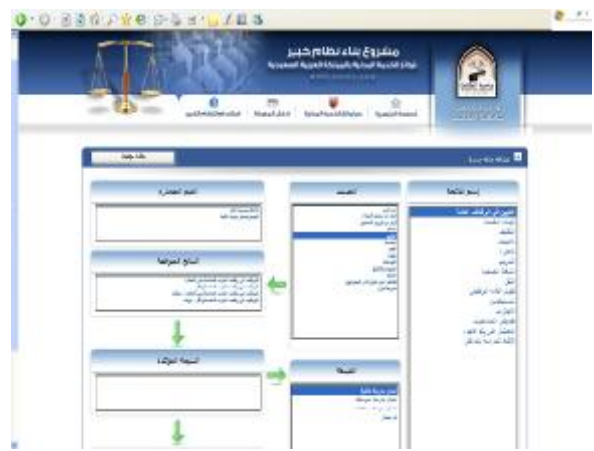

Figure 10. (c): Real example using RCSES step 3.

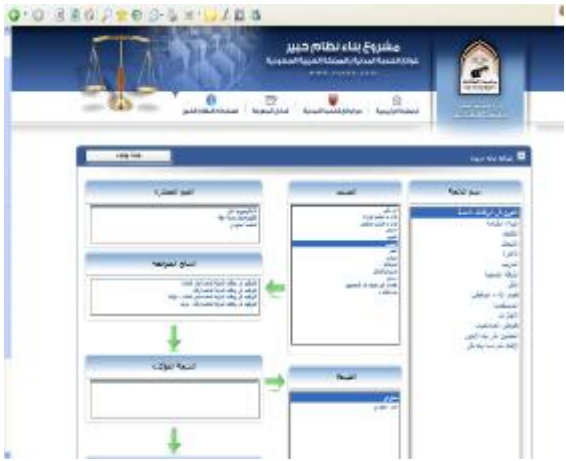

Figure 10. (d): Real example using RCSES step 4.

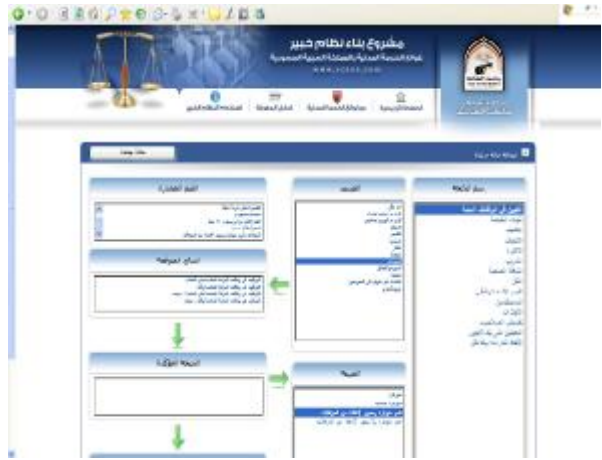

Figure 10. (g): Real example using RCSES step 7.

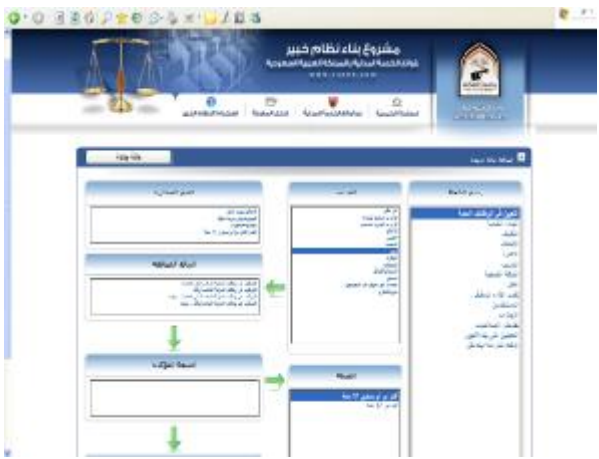

Figure 10. (e): Real example using RCSES step 5.

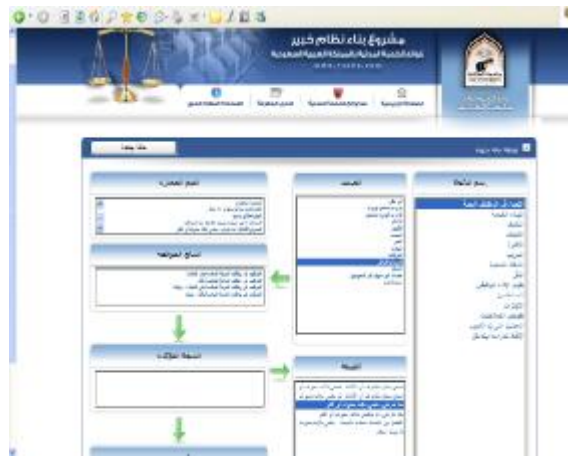

Figure 10. (h): Real example using RCSES step 8.

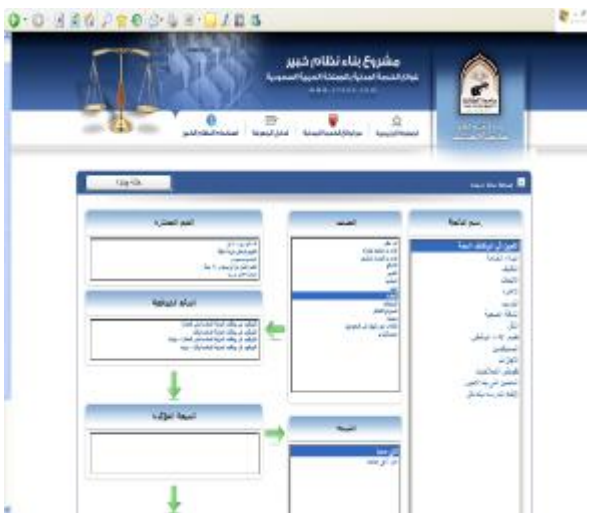

Figure 10. (f): Real example using RCSES step 6.

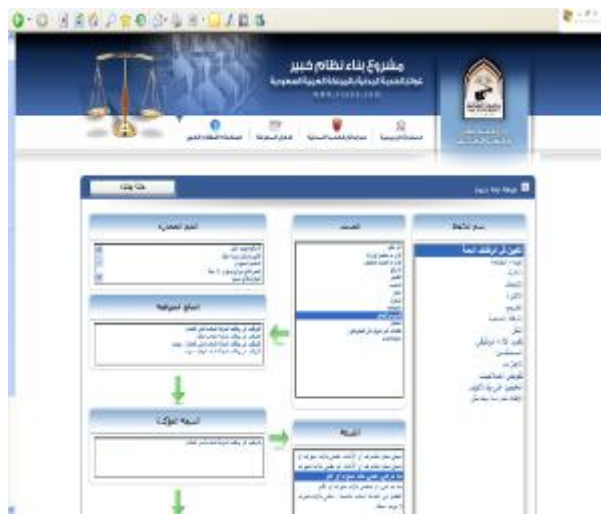

Figure 10. (j): Real example using RCSES step 9.

TABLE I. COMPARISON ANALYSIS WITH 6 RELATED WEB BASED EXPERT SYSTEMS.

| KR | KBR | KM | KB | IE | EP | PE | RT to show result |
|---|---|---|---|---|---|---|---|
| *****The proposed presented work: RCSES***** | | | | | | | |
| Rules were built in XML format | XML | Very easy to modify or change as required | By a developed tool- Knowledge Builder) | backward chaining ,depends on XML parser | Web-Based | .NET technology | Fast |
| **1. Web-based weather expert system (WES) for Space Shuttle launch [16]** | | | | | | | |
| Rules are built in with Java program. | Boolean object | Difficult, by programmer | built in Java program | Backward chaining technique | Web-Based | Java servlet program | Need to download data as inputs |
| **2. Web-based expert system for advising on herbicide use in Great Britain [17]** | | | | | | | |
| Prolog KB | Prolog KB files | Difficult, By programmer | Prolog | Prolog-based inference engine | Web-Based | Perl, and HTML | Takes more time (reads KB files & construct HTML |
| **3. Dr. Wheat: A Web-based Expert System for Diagnosis of Diseases and Pests in Pakistani Wheat [18]** | | | | | | | |
| Form of IF-THEN rules | e2gLite knowledge base | Little difficult, using e2gLite shell | e2gLite™ expert system shell | Forward and backward chaining technique | Web-Based | JAVA | Rather fast, because interface is built by java that processes the built knowledge |
| **4. A Novel Web based Expert System Architecture for On-line and Off-line Fault Diagnosis and Control (FDC) of Power System Equipment [19]** | | | | | | | |
| Form of production rules, | Text files for if-else rules. | Upload the Knowledge base files in text format | Developed Shell | is developed using VB.NET | Web-Based | .NET technology | Rather slow because it depends on saving data in data base and upload the knowledge base after each updating |
| **5. Web-based expert system for food dryer selection [20]** | | | | | | | |
| Rule-based knowledge | Depends on three modules | By Developed ReSolver | ReSolver KB | backward chaining, by ReSolver inference mechanism | Web-Based | Java servlet technology | Takes more times ,create the HTML forms for presenting intermediate and final results and help screens |
| **6. Knowledge Representation and Reasoning Using XML [21]** | | | | | | | |
| using the XML | XML | Very hard as building from start | As text file represents XML format | backward chaining ,Depends on XML parser | Client-side | Visual basic 6.0 with XML | reducing the response time |

web-based ES on the civilian service regulations on the web. There also appears to be a lack of a general methodology for developing web-based expert systems. The proposed system provides better performance as the KR was done by XML format as well as the XML ontology format, the homogeneity between the KR and the platform on the WEB so the performance is very high compared with other systems that needs a mediator to match between the knowledge representation methods and the web platform representation or to load the knowledge from any file or by executing a programs as presented in the comparison. The dynamic property for changing and modifying the Knowledge as needed in a fast and very easy way.

## VII. CONCLUSION

The work presented in this paper tries to overcome the general lack of research in the area of web-based expert systems (WBES). The paper addressed the issues associated with the analysis, design, development, and use of web-based expert system for the regulations of the civil service system in the K.S.A "RCSES". It is the first time to develop such system; which is web based, with the new methodology, the ontology, the knowledge representation, and the tools as well. The presented work introduces a comparison with other related WBES. The work considered 17 regulations for the civil service and its ability to modify or updating and the extending of the existing regulations. The "RCSES" was verified, and validated by different technical users and the developers as well to be usable in the real world governmental departments. The research provides benefits to the employees and the ability to solve the contradiction in the confused problems, as well as the providing for the suggested solutions. The developed RCSES is fully implemented to run on the web using ASP.net techniques as the main programming language, and a new server-side technology, XML-Rule-based knowledge sources and the inference mechanisms were implemented using ASP.net. The work presented in this paper can be extended to add another version of the system in English language, and to include the uncertainty using the fuzzy knowledge representation and inference too.

## Acknowledgment

The authors wish to thank Taif University for the financial support of this research and the anonymous reviewers for their valuable comments.

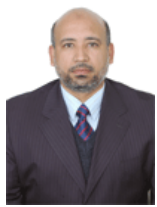

**Mofreh A. Hogo** is a lecturer at Benha University, Egypt. He is a lecturer of Computer Science and Engineering. Dr. Hogo holds a PhD in Informatics Technology from Czech Technical University in Prague, Computer Science and Engineering Dept. 2004. He is the author of over 40 papers that have been published in refereed international Journals  (Information Sciences, Elsiver, UBICC, IJICIS, IJCSIS, IJPRAI, ESWA, IJEL, Web Intelligence and Agent Systems, Intelligent Systems, international journal of NNW, IJAIT Journal of Artificial Intelligence Tools, IJCI) and Book chapters (Neural Networks Applications in Information Technology and Web Engineering Book, Encyclopedia of Data Warehousing and Mining, and Lecture Notes in Artificial Intelligence Series), and international conferences (Systemics, Cybernetics and Informatics Information Systems Management, IEEE/WIC, IEEE/WIC/ACM, ICEIS). His areas of interest include Digital Image Processing, Multimedia Networks, Intrusion detection, Data Mining, Data Clustering and classification, pattern Recognition, character recognition, fuzzy clustering, artificial Neural Networks, Expert systems, Software Engineering.

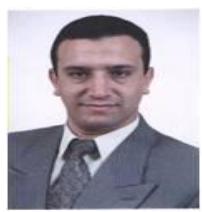

**Khaled Fouad**  is an assistant lecturer at Central Lab. for Agricultural Expert Systems (CLAES), as he working in his Ph.D program. He interested in Programming,  Web  design,  Internet  and  its applications Development, and Expert Systems applications Development.

**Fouad Mousa** is associate professor at Business management Dept,Faculty of commerce, Assuit university.